\documentclass[conference,a4paper]{IEEEtran}
\IEEEoverridecommandlockouts

\usepackage[hidelinks]{hyperref}
\usepackage[cmex10]{amsmath}%American Math Society(AMS) math formatting
\usepackage{amssymb,amsfonts}%AMS extra symbols and fonts
\usepackage[ruled,vlined,linesnumbered,algo2e]{algorithm2e}
\usepackage{dblfloatfix}%fix double column figure ordering and placement
\usepackage{multirow}
\usepackage{subfloat}
\usepackage[ruled,vlined]{algorithm2e}
\usepackage{graphicx}
\graphicspath{{Figures/PDF/}{Figures/PNG/}}
\usepackage{caption}
\usepackage{graphicx}
\usepackage{subfig}
\usepackage{subcaption}
\usepackage{algorithm}
\usepackage{algpseudocode}
\usepackage{booktabs}
\usepackage{siunitx}
\usepackage[numbers,compress]{natbib}
\usepackage{texnames}
\usepackage{bm,bbm}
\usepackage{orcidlink}
\usepackage{balance}

\makeatletter
\usepackage[a4paper, top=1.6cm, bottom=1.6cm, left=1.1cm, right=1.1cm]{geometry}
\makeatother

\begin{document}

\title{CP2M: Clustered-Patch-Mixed Mosaic Augmentation for Aerial Image Segmentation%
\thanks{This research was conducted with the financial support of Science Foundation Ireland under Grant Agreement No.\ 13/RC/2106\_P2 at the ADAPT SFI Research Centre at University College Dublin. The ADAPT Centre for Digital Content Technology is partially supported by the SFI Research Centres Programme (Grant 13/RC/2106\_P2) and is co-funded under the European Regional Development Fund.}%
\thanks{Send correspondence to \mbox{S.\ Dev, E-mail: soumyabrata.dev@ucd.ie.}}}

\author{%
    \IEEEauthorblockN{Yijie~Li\orcidlink{0000-0003-2118-2280}$^{1}$, Hewei~Wang\orcidlink{0000-0002-6952-0886}$^{2}$, Jinfeng~Xu\orcidlink{0009-0001-7876-3740}$^{3}$, Zixiao~Ma\orcidlink{0009-0001-4587-0128}$^{4}$, Puzhen~Wu\orcidlink{0000-0003-1510-215X}$^{5}$, Shaofan~Wang\orcidlink{0000-0002-3045-624X}$^{6}$, Soumyabrata~Dev\orcidlink{0000-0002-0153-1095}$^{5,7}$}
    \IEEEauthorblockA{%
        $^{1}$Northwestern University
        $^{2}$Carnegie Mellon University
        $^{3}$The University of Hong Kong  
        $^{4}$Duke University \\
        $^{5}$University College Dublin
        $^{6}$Beijing University of Technology
        $^{7}$The ADAPT SFI Research Centre
    }
}

\maketitle
\begin{abstract}
Remote sensing image segmentation is pivotal for earth observation, underpinning applications such as environmental monitoring and urban planning. Due to the limited annotation data available in remote sensing images, numerous studies have focused on data augmentation as a means to alleviate overfitting in deep learning networks. However, some existing data augmentation strategies rely on simple transformations that may not sufficiently enhance data diversity or model generalization capabilities. This paper proposes a novel augmentation strategy, Clustered-Patch-Mixed Mosaic (CP2M), designed to address these limitations. CP2M integrates a Mosaic augmentation phase with a clustered patch mix phase. The former stage constructs a new sample from four random samples, while the latter phase uses the connected component labeling algorithm to ensure the augmented data maintains spatial coherence and avoids introducing irrelevant semantics when pasting random patches. Our experiments on the ISPRS Potsdam dataset demonstrate that CP2M substantially mitigates overfitting, setting new benchmarks for segmentation accuracy and model robustness in remote sensing tasks. In the spirit of reproducible research, the code, dataset, and experimental results are publicly available at: \url{https://github.com/Att100/CP2M}.
\end{abstract}

\section{Introduction}
Remote sensing technology, through its capacity to extensively survey and analyze the Earth's surface, plays a crucial role across various domains, including climate change mitigation \cite{gordon2023remote}, biodiversity conservation \cite{teng2023bird}, meteorological analysis \cite{li2024ucloudnet}, rainfall prediction \cite{wang2021day}, and energy management \cite{gaumer2022estimating}. This field's evolution has been significantly driven by deep learning, which has enhanced the accuracy and efficiency of analyzing satellite imagery for tasks such as monitoring greenhouse gas emissions from power plants \cite{scheibenreif2021estimation},  estimating energy needs in remote regions \cite{jain2022employing}, superpixel-based hyperspectral image clustering \cite{cui2024superpixel}, and real-time localization of vegetation patterns using UAV imagery \cite{cui2024real}. The effectiveness of these approaches is underpinned by the growing availability and complexity of geospatial datasets, enabling detailed assessments of environmental impacts and aiding in the transition towards renewable energy sources \cite{lynch2021leveraging} and conservation efforts \cite{teng2023bird}. Furthermore, the development of standardized evaluation platforms like the GRSS Data and proposal for the Climate Change Benchmark \cite{lacoste2021toward} underscores the field's progression towards more universal and accessible remote sensing applications.

However, the application of deep learning in remote sensing faces overfitting challenges led by the high dimensionality of data and the complexity of spatial patterns. Image augmentation has emerged as a pivotal technique in addressing these challenges, effectively expanding the diversity of training datasets and improving model robustness. Despite these advances, models often suffer from overfitting due to the limited variability within the augmented datasets. Recognizing these issues, recent efforts have focused on developing data augmentation \cite{doi:10.1080/15481603.2017.1323377} and regularization techniques \cite{devries2017improved, yun2019cutmix} to bolster model performance and versatility. 
%2020predicting, prapas2022deep
To ameliorate the aforementioned issues, we propose the CP2M model, a novel approach designed to enhance model generalization and mitigate overfitting in remote sensing applications.

\begin{figure*}[htbp]
    \centering
    \scalebox{0.52}{
        \includegraphics{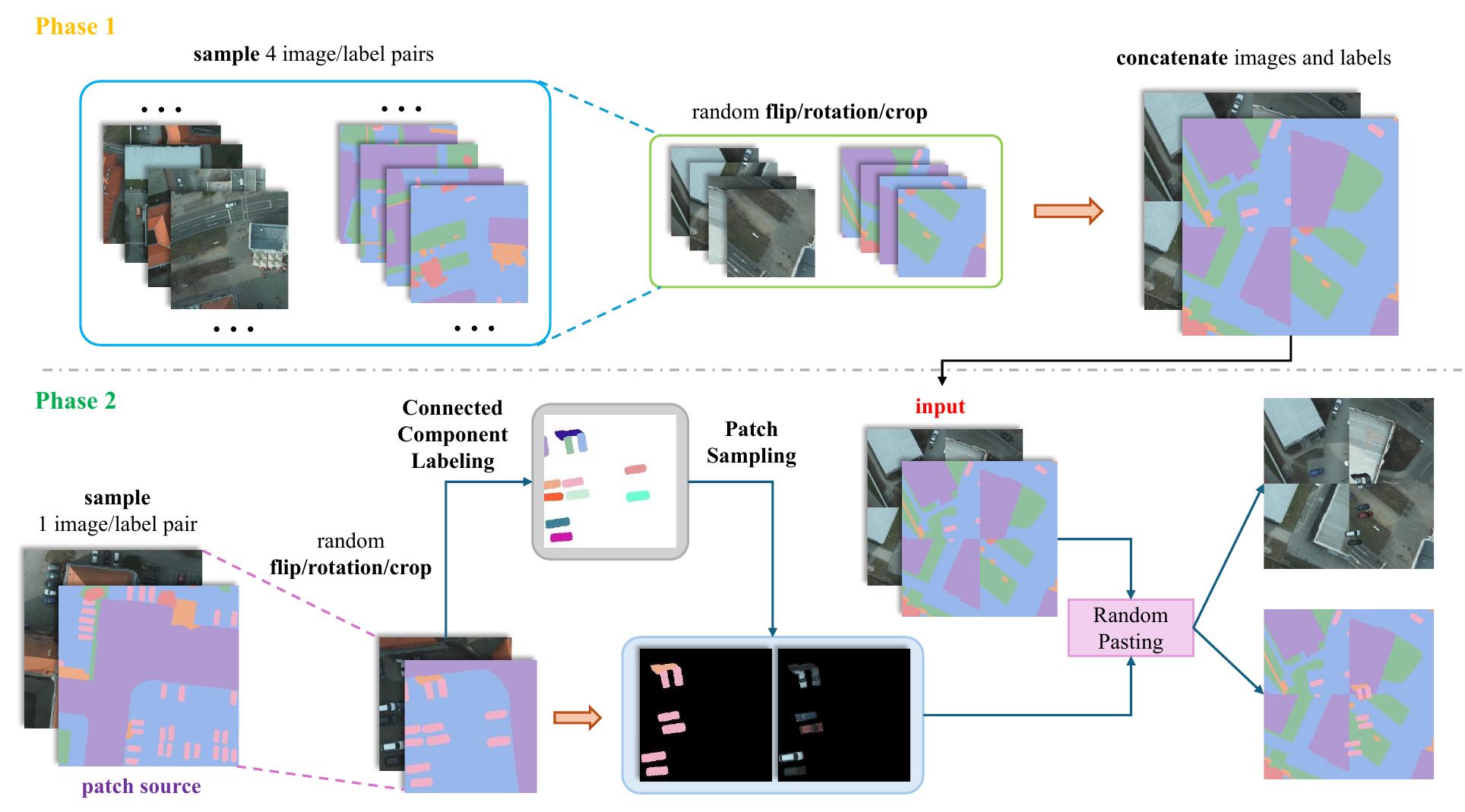}
    }
    \caption{Overall pipeline of Clustered-Patch-Mixed Mosaic (CP2M)}
    \label{fig:cp2m-pipeline}
    \vspace{-0.6cm}
\end{figure*}

\section{Related Work}
In modern aerial image segmentation, deep learning-based approaches have become mainstream. In 2025, Long \textit{et al.} proposed the FCN \cite{long2015fully} model for semantic segmentation, which lays the foundation for semantic segmentation. In the same year, Ronneberger \textit{et al.} proposed the U-Net \cite{ronneberger2015u} for medical image segmentation which became the most widely-used base model for aerial image segmentation. In order to enable the mobile-device inference ability, Chen \textit{et al.} proposed DeepLab V3+ \cite{chen2018encoder} in 2018, utilizing the Atrous Separable Convolution for efficient image segmentation. After the introduction of the Transformer \cite{vaswani2017attention} and ViT (vision transformer) \cite{dosovitskiy2020image}, the performance of aerial image segmentation is boosted. In 2024, Yamazaki \textit{et al.} proposed the AerialFormer \cite{hanyu2024aerialformer}, a specially designed Transformer-based model for accurate aerial image segmentation. However, labeling aerial images is time-consuming and expensive. Data augmentation has become an indispensable technique in improving the performance of deep learning models for remote sensing image analysis while reducing the data labeling costs. % Hao et al. \cite{rs15030827} conducted a comprehensive review of data augmentation methods specifically designed for remote sensing image target recognition. Their work highlighted traditional methods and recent advancements, offering insights into strategies to address the unique challenges of remote sensing datasets, such as class imbalance and limited labeled data. 
Lu et al. \cite{rsi2022} proposed RSI-Mix, a novel data augmentation technique for remote sensing image classification. By integrating region-specific mixing strategies, RSI-Mix demonstrated significant improvements in classification performance, particularly in scenarios with limited training data. Khammari et al. \cite{synthetic2024} explored synthetic data generation for earth observation object detection tasks. Their method focused on embedding synthetic objects into satellite imagery, creating diverse and realistic datasets to enhance model training. Tang et al. \cite{aerogen2024} introduced AeroGen, a diffusion-driven data generation method tailored for remote sensing object detection. This approach effectively addresses the scarcity of annotated data by synthesizing realistic variations of objects, thereby boosting detection accuracy. %In 2024, Karimli et al. \cite{change2024} investigated data augmentation in the context of remote sensing image change captioning. %Their approach tackled the challenge of generating captions for land-cover transformations by employing advanced augmentation techniques to increase dataset variability and improve model generalization.

% These studies collectively emphasize the critical role of data augmentation in remote sensing, ranging from traditional techniques to cutting-edge generative methods, paving the way for more robust and accurate models in remote sensing domain.

\section{CP2M Pipeline}
Our proposed CP2M augmentation is shown in Figure \ref{fig:cp2m-pipeline} which consists of two phases, Enhanced Mosaic augmentation and Clustered Patch Mix (CPM) augmentation. In our approach, there are two thresholds for controlling the probability of applying Mosaic and CPM augmentation.

\subsection{Enhanced Mosaic Augmentation}
Mosaic Augmentation was first introduced by Bochkovskiy \textit{et al.} in YOLOv4 for object detection \cite{bochkovskiy2020yolov4}. The Mosaic augmentation mixes four different images that enable the model learning to detect objects outside their normal context. Another advantage of Mosaic is combining four images into one can increase the number of samples in one batch without increasing batch size. In semantic segmentation, particularly in specialized domains such as autonomous driving and medical imaging, merging images could significantly distort the context and feature distribution of the original images. However, in semantic segmentation for remote sensing, the variability in the positional distribution of object classes within images renders the technique less sensitive to the contextual disruptions introduced by Mosaic, avoiding performance decline from stitching multiple images. Furthermore, given the high resolution of remote sensing images and the extensive costs associated with data labeling, data augmentation is essential for improving segmentation performance in this field.

As shown in Figure \ref{fig:cp2m-pipeline} Phase 1, we first sample four different images from the training set. For each of these images, we apply random vertical/horizontal flip, rotation, and crop on both the RGB image and its corresponding label. Subsequently, we concatenate the four processed samples together by positioning them in the four quadrants of a new image canvas. To guarantee the consistency of resolution between the Mosaic augmented sample and the original samples, we randomly crop each sub-image to half the size of the assembled image. During sample loading, a random number between 0 and 1 is generated, using as a threshold $\mathrm{p-mosaic}$ (representing the probability of using Mosaic augmentation) to control the proportion of normal and Mosaic samples in the training stage.

\begin{figure}[htbp]
    \centering
    \scalebox{0.3}{
        \includegraphics{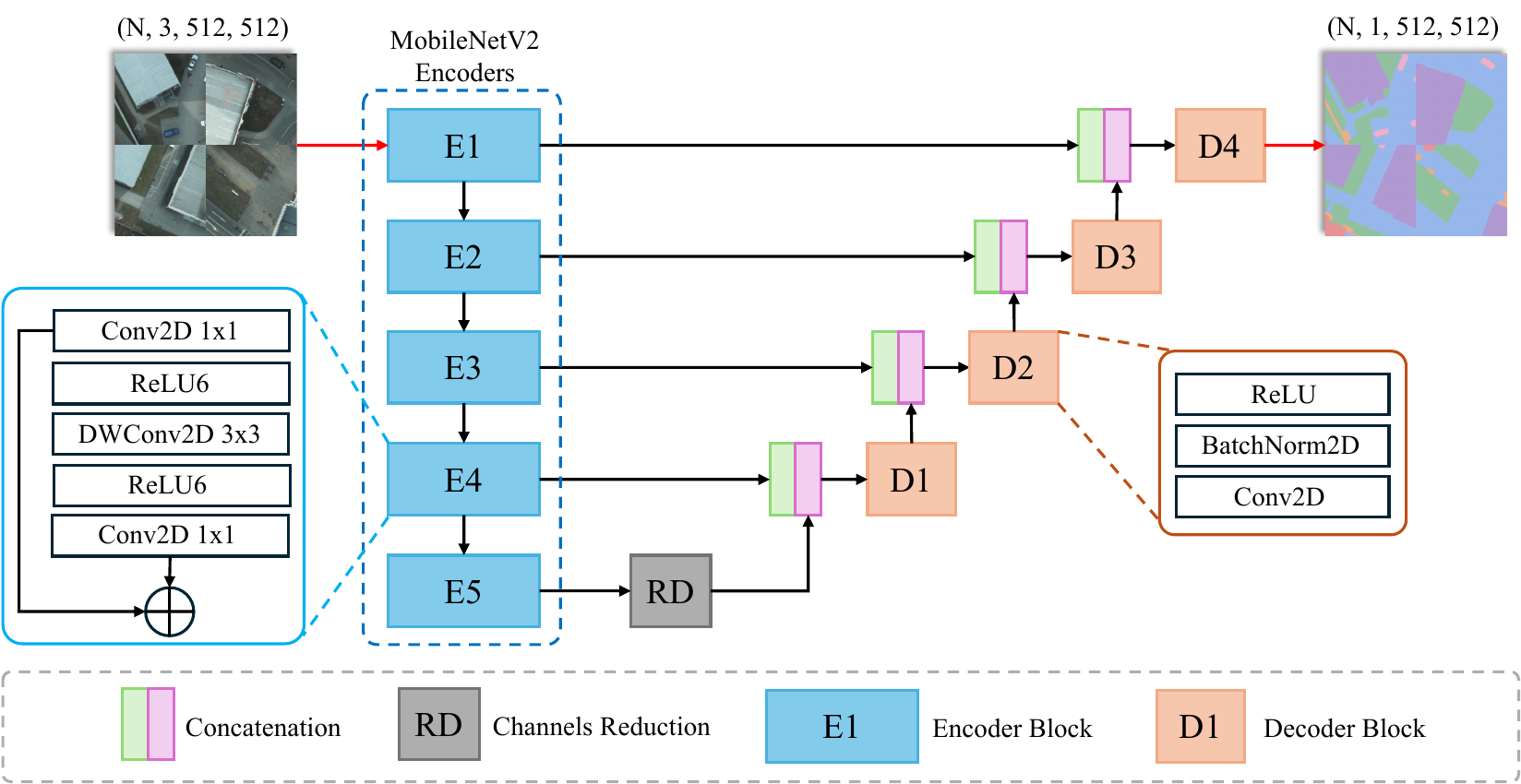}
    }
    \caption{MobileNetV2-UNet Model Architecture}
    \label{fig:cp2m-unet}
\end{figure}

\subsection{Clustered Patch Mixed Augmentation}
Yun \textit{et al.} \cite{yun2019cutmix} proposed a regularization strategy called CutMix whose core idea is to create a new training image by combining two different images. This is done by cutting and pasting rectangular patches from one image to another. Specifically, a patch from one image is cut and pasted onto another image. The labels of these images are also mixed proportionally to the area of the patches. However, the CutMix approach pastes both target and irrelevant pixels which may have a negative impact on semantic segmentation which motivated us to introduce the Clustered Patch Mixed (CPM) Augmentation, shown in Figure \ref{fig:cp2m-pipeline} Phase 2.

Since objects of the same class, especially trees and vehicles, are presented as scattered and disconnected clusters in remote sensing images, this allows us to apply the connected component labeling algorithm to separate objects of the same class in the form of different instances, which we call each instance a patch. We first sample an image/label pair from the training set, we apply random horizontal and vertical flips, rotation, and crop to both the image and the label. Then for each class in the label, we run the connected component labeling algorithm to separate disjointed objects into different instances and label them with different IDs. 

Connected Component Labeling (CCL) is an algorithm employed in computer vision to identify connected regions within binary images, which consist exclusively of pixels valued `0' or `1'. The objective of CCL is to discern and assign a unique label to each connected component among the foreground pixels, defined as a cluster of adjacent pixels where each is directly neighboring at least one other within the same cluster. In our CP2M pipeline, we randomly sample $k$ different instances and paste them to the input image and label, which can be formulated as:
\begin{align}
  \left\{
    \begin{aligned}
      \hat{\mathrm{image}} &= \mathrm{image} \odot (1 - \mathrm{mask}) + \mathrm{image}^{\prime} \odot \mathrm{mask}, \\
      \hat{\mathrm{label}} &= \mathrm{label} \odot (1 - \mathrm{mask}) + \mathrm{label}^{\prime} \odot \mathrm{mask}
    \end{aligned}
  \right.
\end{align}
where $\mathrm{image}$ and $\mathrm{label}$ are input image and corresponding label. $\mathrm{image}^{\prime}$ and $\mathrm{label}^{\prime}$ are image and label of patch source. $\hat{\mathrm{image}}$ and $\hat{\mathrm{label}}$ are output image and label. $\mathrm{mask}$ represents the binary mask indicating selected pixels. $\odot$ is element-wise multiplication.

\begin{figure}[htbp]
    \centering
    \scalebox{0.26}{
        \includegraphics{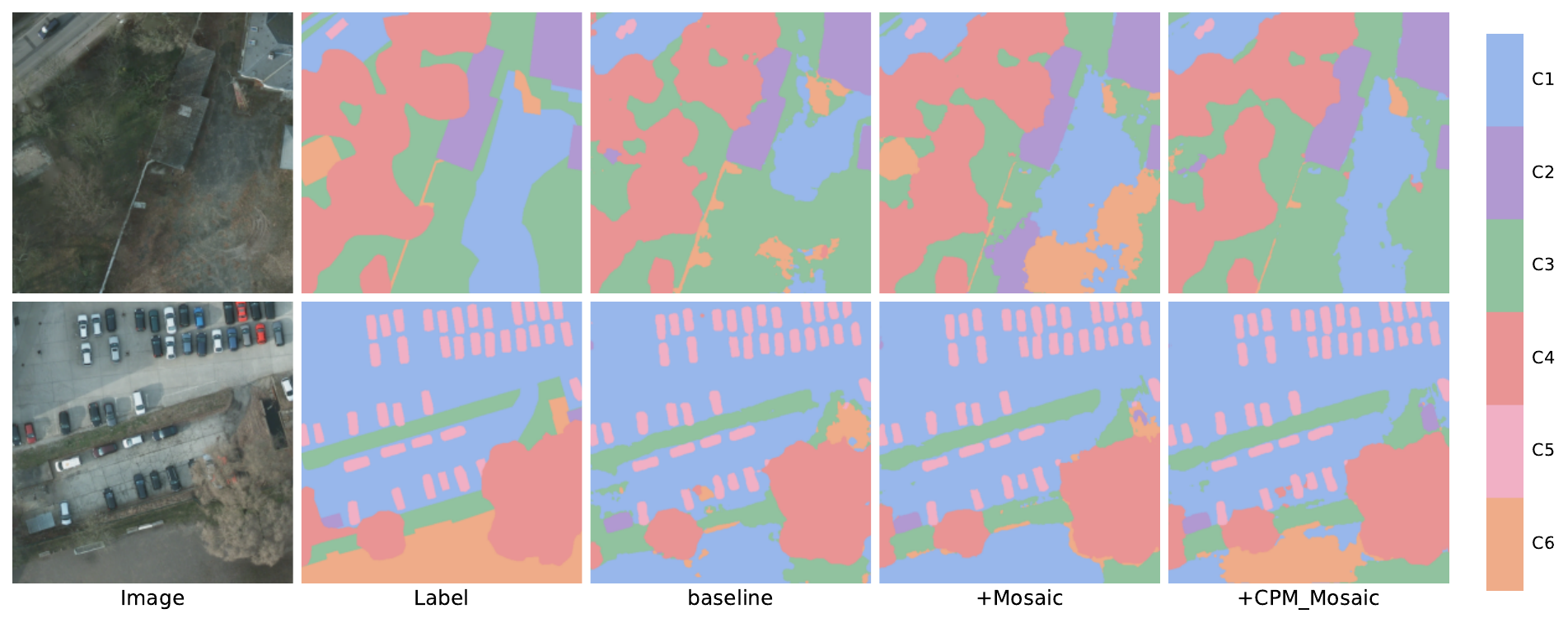}
    }
    \caption{Qualitative comparison between baseline and CP2M. class mapping: impervious surfaces (C1), buildings (C2), low vegetation (C3), trees (C4), cars (C5), clutter/background (C6).}
    \label{fig:qualitative}
\end{figure}

\section{Experiments \& Results}
\subsection{Experiments Settings}
\label{sec:exp-set}
Our experiments are conducted on the Potsdam dataset \cite{isprs2024potsdam}, which contains 38 image/label pairs with $6000 \times 6000$ resolution. We follow the official settings, using 24 samples for training and 14 samples for testing. We use a $1000 \times 1000$ sliding window to get 864 samples for training and 504 samples for testing. We use the U-Net \cite{ronneberger2015u} with MobileNetV2 \cite{sandler2018mobilenetv2} backbone as the segmentation model. We use Adam optimizer with a constant learning rate of 1e-4, and L2 weight-decay of 4e-5. The batch size is 8 and all experiments are trained for 50 epochs. 

\subsection{Model Architecture and Objective Function}

The UNet-based \cite{ronneberger2015u} model architecture used in our experiments is shown in Figure \ref{fig:cp2m-unet}, which contains a MobileNetV2 \cite{sandler2018mobilenetv2} encoder, a channels reduction module, and four decoders. Encoder-decoder architectures are commonly employed to process 2D image data via CNNs~\cite{WANG2022102243, WANG2021, BATRA2022200039}. This kind of model architecture finds extensive applications in diverse fields including robotics~\cite{zhenqi23, HuoAtten, zhu2023fanuc, wang2024airshot, wang2024onls}, self-driving perception~\cite{WANGCAR2022}, salient object detection~\cite{li2023daanet, Li_2024_BMVC}, medical vision~\cite{tang2024optimized, pan2024accurate}, recommendation systems~\cite{xu2024aligngroup, xu2024mentor, xu2024fourierkan}. For our model architecture, the channel reduction module is a group of a 2D convolution layer, a 2D batch-normalization layer, and a ReLU activation. Each decoder consists of two groups of 2D convolution, batch-normalization, and ReLU. We use an extended cross entropy as the training objective function which is defined as:

$$
\mathcal{L}_{ce} = - \sum_{i=1}^{C} w_{i} y_{i}\log(p_{i}) + \lambda \mathcal{R}(\theta)
$$
where $y_{i}$ and $p_{i}$ represent the label and predicted probability for the $i$-th class, $w_{i}$ is the class weight to address class imbalance, $\lambda$ controls the regularization strength, and $\mathcal{R}(\theta)$ (e.g., L1/L2) prevents overfitting by penalizing large weights.

\begin{figure}[htbp]
    \centering
    \scalebox{0.26}{\includegraphics{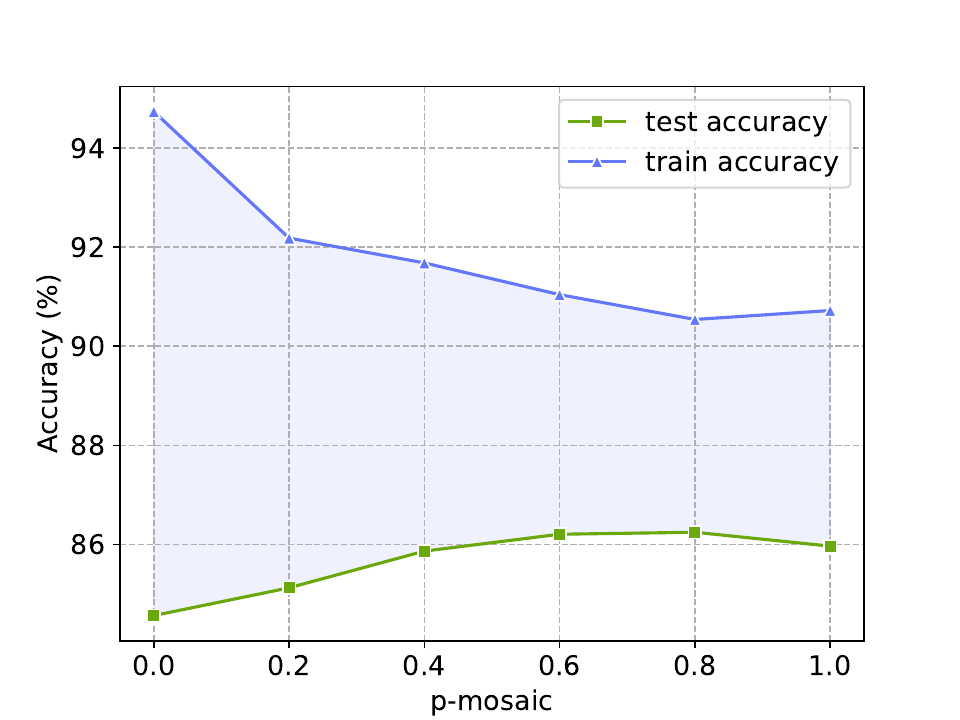}}
    \scalebox{0.26}{\includegraphics{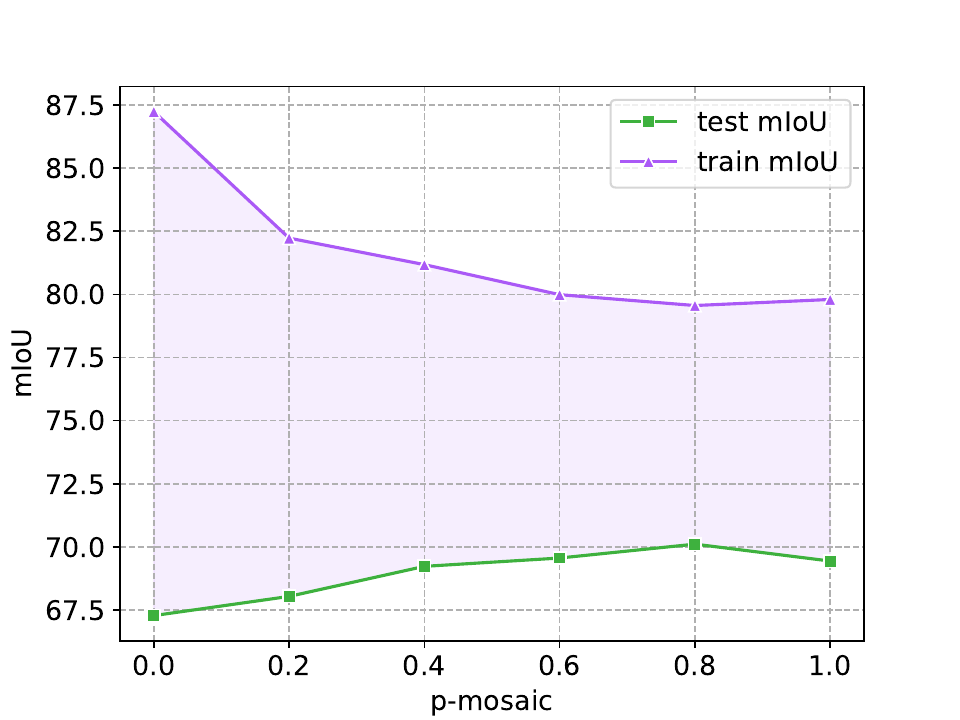}}
    \caption{Impact of the probability of using Mosaic on model performance (Accuracy and mIoU). We use $\mathrm{p\_mosaic}$ as a threshold to control the proportion of normal and Mosaic samples in training. $\mathrm{p\_mosaic}$ is 0 means Mosaic is not used, $\mathrm{p\_mosaic}$ is 1 means Mosaic is used for every sample.}
    \label{fig:p_mosaic}
    \vspace{-0.6cm}
\end{figure}

\subsection{Metrics}
 We quantitatively assessed using mIoU and accuracy metrics. The mIoU is a class-wise averaged metric, formally defined for each class $c$ as the ratio of the true positive predictions $TP_c$ to the sum of true positive, false positive $FP_c$, and false negative $FN_c$ predictions:
\begin{equation}
\text{mIoU} = \frac{1}{N_{\text{classes}}} \sum_{c=1}^{N_{\text{classes}}} \frac{TP_c}{TP_c + FP_c + FN_c},
\end{equation}

\subsection{Qualitative Evaluation}
Figure \ref{fig:qualitative} visually compares baseline segmentation results with those using CP2M augmentation. Column 1 shows the original images, while column 2 displays ground truth labels for classes such as impervious surfaces (C1), buildings (C2), low vegetation (C3), trees (C4), cars (C5), and clutter/background (C6). Column 3 (baseline predictions) moderately aligns with the ground truth but shows notable misclassifications, particularly between C3 and C4 and in distinguishing C5 from its surroundings. Mosaic augmentation (column 4) enhances accuracy, improving the delineation of C2 and detection of C5. The final column, incorporating both Mosaic and CPM Mosaic augmentations, shows further improvements, with finer segmentation of C4, clearer distinction of C3, and fewer misclassifications in C6.

% Figure \ref{fig:qualitative} presents a visual comparison between the baseline segmentation results and those obtained using the CP2M augmentation. The first column displays the original images, while the second column shows the ground truth labels for various classes such as impervious surfaces (C1), buildings (C2), low vegetation (C3), trees (C4), cars (C5), and clutter/background (C6). The third column, representing the baseline model's predictions, shows moderate alignment with the ground truth but exhibits notable misclassifications, particularly in distinguishing between C3 and C4, as well as differentiating C5 from their surroundings. The addition of Mosaic augmentation, as seen in the fourth column, enhances the model's accuracy, with a more precise delineation of building C2 and improved detection of individual C5. The fifth column, which incorporates both Mosaic and CPM Mosaic augmentations, reveals further improvements. Here, we observe a finer segmentation of the C4, a better distinction of C3, and fewer misclassified regions in C6.

%These qualitative observations suggest that CP2M significantly refines the segmentation task, contributing to more accurate and granular interpretations of complex urban landscapes in remote sensing imagery.

\begin{figure}[htbp]
    \centering
    \scalebox{0.20}{
        \includegraphics{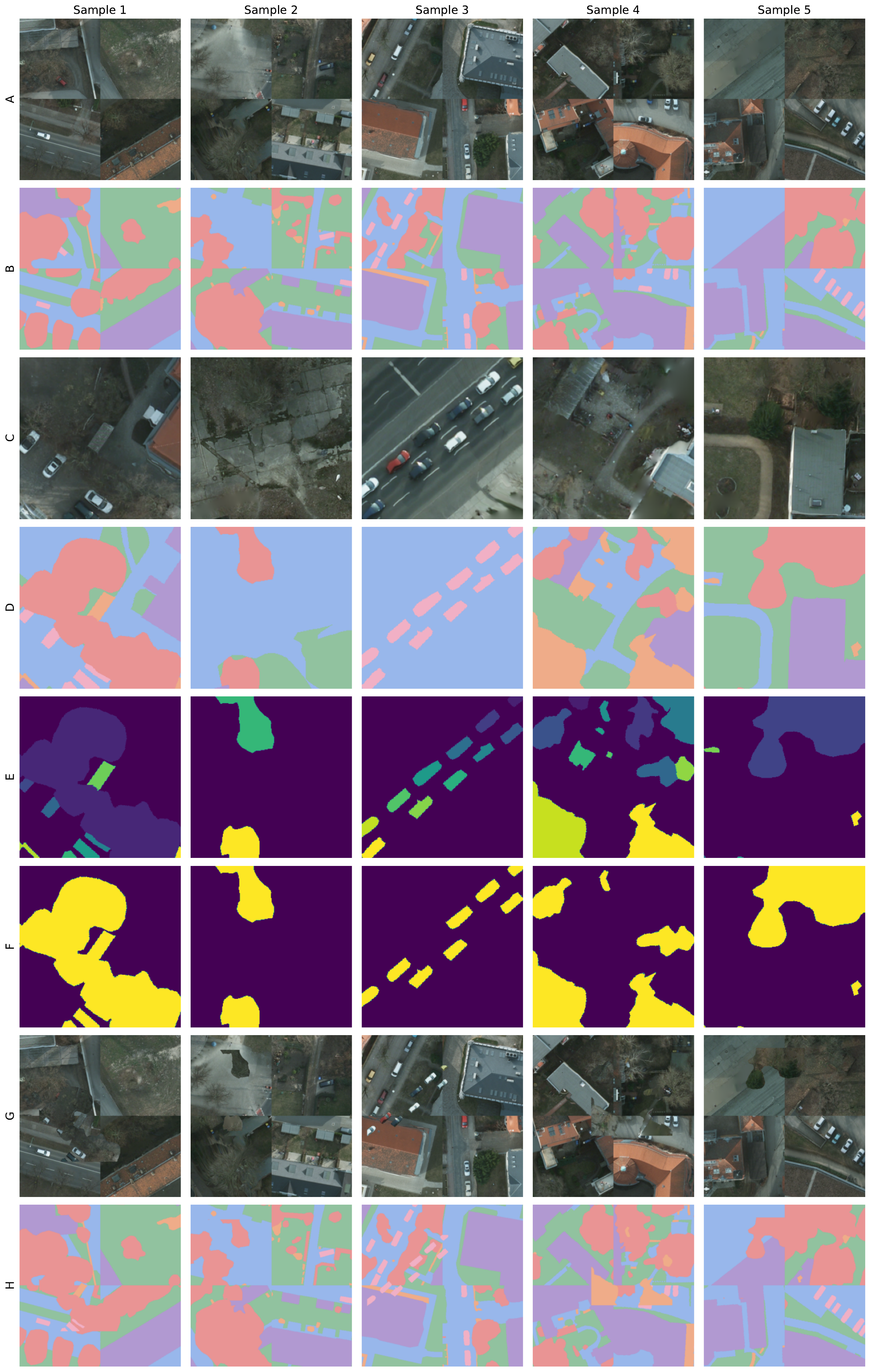}
    }
    \caption{CP2M Augmentation Process: (A) Original Mosaic Images, (B) Mosaic Labels, (C) Patch Images, (D) Patch Labels, (E) Connected Component Labeling Results with Color-coded Instances, (F) Binary Masks Selection, (G) Augmented Output Images, 
    (H) Final Augmented Output Labels.}
    \label{fig:cp2m-vis2}
    \vspace{-0.6cm}
\end{figure}

\subsection{Quantitative Evaluation}

\begin{table}[htbp]
  \centering
  \caption{Quantitative Results of CP2M}
  \label{tab:quant}
  \setlength\tabcolsep{4pt} % Adjust the column separation
  \scalebox{0.81}{
      \begin{tabular}{@{}l|S[table-format=2.2,table-space-text-post=\textcolor{red}{\textbf{ (+3.25)}}]S[table-format=2.2,table-space-text-post=\textcolor{red}{\textbf{ (+3.25)}}]SSSSSS@{}}
          \toprule
          Method & {Accuracy} & {mIoU} & {C1} & {C2} & {C3} & {C4} & {C5} & {C6} \\
          \midrule
          Baseline & 84.57 & 67.29 & 78.31 & 86.87 & 67.15 & 67.26 & 76.34 & 27.74 \\
          + Mosaic & 86.25  & 70.11  & 80.46 & 88.32 & 69.00 & \textbf{71.44} & \textbf{77.79} & 33.65 \\
          + CP2M & \textbf{86.74} \textcolor{red}{\textbf{ (+2.17)}} & \textbf{70.54} \textcolor{red}{\textbf{ (+3.25)}} & \textbf{81.16} & \textbf{89.05} & \textbf{70.60} & 71.17 & 77.03 & \textbf{34.23} \\
          \bottomrule
      \end{tabular}
  }
\end{table}

The quantitative evaluation in Table \ref{tab:quant} highlights the significant effectiveness of the CP2M model. By integrating CP2M, notable improvements are observed in metrics such as accuracy and mIoU compared to the baseline. Mosaic augmentation alone increases accuracy from 84.57\% to 86.25\% (+1.68\%) and mIoU from 67.29\% to 70.11\% (+2.82\%), showcasing its impact. Adding CPM Mosaic further boosts accuracy by 2.17\% and mIoU by 3.25\% over the baseline. These enhancements starkly highlight the CP2M model's capability in mitigating overfitting and boosting segmentation accuracy for remote sensing tasks. We also measure the per-class IoU which witness significant performance gains in most classes. Figure \ref{fig:p_mosaic} demonstrates the impact of the probability of using Mosaic on accuracy and mIoU during training/test, revealing that 100\% Mosaic data augmentation strategy is suboptimal.

% The quantitative evaluation of the CP2M model, as shown in Table \ref{tab:quant}, illustrates its significant effectiveness. By integrating CP2M, notable improvements are observed across critical metrics such as accuracy and mIoU compared to the baseline model. Specifically, the application of Mosaic augmentation alone increases overall accuracy from 84.57\% to 86.25\%, marking a gain of approximately 1.68\%. This augmentation also raises the mIoU from 67.29\% to 70.11\%, a boost of 2.82\%, underscoring its substantial impact on model performance. The incorporation of CPM Mosaic augmentation further lifts overall accuracy from the baseline with an improvement of 2.17\%. Likewise, mIoU sees a rise with a significant uplift of 3.25\%. These enhancements starkly highlight the CP2M model's capability in mitigating overfitting and boosting segmentation accuracy for remote sensing tasks. We also measure the per-class IoU which witness significant performance gains in most classes. Figure \ref{fig:p_mosaic} demonstrates the impact of the probability of using Mosaic on accuracy and mIoU during training/test, revealing that 100\% Mosaic data augmentation strategy is suboptimal.

\subsection{Visualization of CP2M Generated Samples}
% In Figure \ref{fig:cp2m-vis2}, we provide five CP2M augmented samples including selected middle outputs. Row A and Row B show the mosaic augmented images and labels. Row C and Row D are the image and label of the patch source mentioned in Figure \ref{fig:cp2m-pipeline}. Images in Row E are results generated via the connected component labeling algorithm and instances are represented by different colors. The binary masks of Row F are instances randomly selected from Row E. The last two rows, G and H demonstrate the final output image and label of CP2M. By comparing Row A/B and Row G/H, it is obvious that our proposed CP2M can significantly increase the diversity of training samples.

In Figure \ref{fig:cp2m-vis2}, we present five augmented samples created using the CP2M technique and illustrate eight essential steps to obtain the final augmented output label with selected intermediate outputs. Rows A and B display the images and labels augmented with the mosaic technique. Rows C and D depict the source image and label for the patch mentioned in Figure \ref{fig:cp2m-pipeline}. The images in Row E are the outcomes of applying the connected component labeling algorithm, where instances are distinguished by various colors. The binary masks in Row F consist of instances randomly chosen from Row E. The final two rows, G and H, showcase the ultimate output image and label generated by CP2M. A comparison between Rows A/B and G/H clearly demonstrates the significant enhancement in the diversity of training samples achieved through our proposed CP2M approach. 

\section{Conclusion}
% In this work, we introduce a novel cluster-patch-mixed Mosaic augmentation strategy for aerial image segmentation. We propose to use the connected component labeling algorithm to sample patches from a given source image and enhance the input sample by pasting the image/label patch to a random location which further increases the diversity of training data and effectively mitigates over-fitting and improves model performance. We also explore the effect of different proportions of data-enhanced and normal samples during the training stage.我

In this work, we introduce a novel cluster-patch-mixed Mosaic augmentation strategy tailored for aerial image segmentation. We propose employing the connected component labeling algorithm to extract patches from a source image, thereby enriching the input sample by affixing the image/label patch to a random position. This approach significantly broadens the diversity of the training data, effectively counteracts overfitting, and enhances model performance. Additionally, we investigate the impact of varying ratios of data-enhanced to normal samples throughout the training phase. %For future work, we intend to integrate advanced generative techniques, such as diffusion models, to automate the creation of more realistic and context-aware augmented samples. %Additionally, we plan to extend CP2M by incorporating domain-specific augmentation strategies and adaptive threshold mechanisms to dynamically adjust the balance between normal and augmented samples during training.

\balance

% \newpage
\bibliographystyle{IEEEtran.bst}
\bibliography{strings,refs}

\end{document}